\documentclass[letterpaper, 10 pt, conference]{format/ieeeconf}

\usepackage{amsmath}
\usepackage{amsfonts}
\usepackage{amssymb}

\usepackage{amsthm}
\usepackage[ruled,vlined]{algorithm2e}

\usepackage{mathtools}
\usepackage{tabularx}
\usepackage{graphicx}
\usepackage{subfigure}
\usepackage{enumerate}
\usepackage{float}
\usepackage{url}
\usepackage{verbatim}
\usepackage{stfloats}
\usepackage{bbding}
\usepackage{bm}
\usepackage{hyperref}
\usepackage{booktabs}
\usepackage{threeparttable}  
\usepackage{multirow}
\usepackage{pifont}
\usepackage{subcaption}
\usepackage{colortbl}
\usepackage{array}
\usepackage{booktabs}
\usepackage{cite}

\newcolumntype{M}[1]{>{\centering\arraybackslash}m{#1}}

\usepackage[table,xcdraw]{xcolor}
\definecolor{lightgreen}{RGB}{222, 238, 240} 
\definecolor{lightours}{RGB}{235, 242, 238}

\definecolor{compact}{RGB}{120, 32, 110}
\definecolor{primitives}{RGB}{0, 32, 96}
\definecolor{context}{RGB}{96, 189, 63}

\graphicspath{{figures/}}
\IEEEoverridecommandlockouts
\makeatletter
\def\@fnsymbol#1{\ensuremath{\ifcase#1\or \dagger\or \ddagger\or
		\mathsection\or \mathparagraph\or \|\or **\or \dagger\dagger
		\or \ddagger\ddagger \else\@ctrerr\fi}}
\makeatother

\title{{\textbf{Resolving State Ambiguity in Robot Manipulation via Adaptive Working Memory Recoding}}}

\author{
Qingda Hu\textsuperscript{1},
Ziheng Qiu\textsuperscript{1},
Zijun Xu\textsuperscript{1},
Kaizhao Zhang\textsuperscript{1},
Xizhou Bu\textsuperscript{1},
Zuolei Sun\textsuperscript{2},\\
Bo Zhang\textsuperscript{2},
Jieru Zhao\textsuperscript{3},
Zhongxue Gan\textsuperscript{1},
and Wenchao Ding\textsuperscript{1}
\thanks{1 Fudan Academy for Engineering and Technology, Fudan University}
\thanks{2 Westwell}
\thanks{3 Department of Computer Science and Engineering, Shanghai Jiao Tong University,
China}
\thanks{Email:{\tt\small qdhu24@m.fudan.edu.cn, dingwenchao@fudan.edu.cn}}
}

\usepackage{float}
\begin{document}
    \makeatletter
    \let\@oldmaketitle\@maketitle
    \makeatother
\maketitle

\begin{figure*}[!t]
    \centering
    \includegraphics[width=\textwidth]{figures/coarse.png}
    \caption{\textbf{Left:} We illustrate a representative example of state ambiguity and elaborate on its various scenarios. State ambiguity is common in robotic manipulation, and the relevant state transitions do not follow the Markov assumption; therefore, a long history window is essential.
    \textbf{Right:} Existing methods and our method for encoding long history. Analogous to human continuous reasoning, PAM’s inference process is temporally dependent, extending the history window by maintaining a long-term adaptive working memory. The model only needs to encode the current observation at each inference step.}
    \label{fig:coarse}
\end{figure*}

\begin{abstract}
State ambiguity is common in robotic manipulation. Identical observations may correspond to multiple valid behavior trajectories. The visuomotor policy must correctly extract the appropriate types and levels of information from the history to identify the current task phase. 
However, naively extending the history window is computationally expensive and may cause severe overfitting. 
Inspired by the continuous nature of human reasoning
and the recoding of working memory,
we introduce PAM, a novel visuomotor Policy equipped with Adaptive working Memory.
With minimal additional training cost in a two-stage manner, PAM supports a 300-frame history window while maintaining high inference speed. 
Specifically, a hierarchical frame feature extractor yields two distinct representations for motion primitives and temporal disambiguation. 
For compact representation, a context router with range-specific queries is employed to produce compact context features across multiple history lengths. 
And an auxiliary objective of reconstructing historical information is introduced to ensure that the context router acts as an effective bottleneck.
We meticulously design 7 tasks and verify that PAM can handle multiple scenarios of state ambiguity simultaneously.
With a history window of approximately 10 seconds, PAM still supports stable training and maintains inference speeds above 20Hz. The project page is \url{https://tinda24.github.io/pam/}.
\end{abstract}

\section{Introduction}
\label{sec:Introduction}
In recent years, a series of visuomotor policies \cite{ACT,dp,pi0,rt2,octo,openvla} based on behavioral cloning have emerged, enabling end-to-end learning of a wide range of skills. 
Most existing methods\cite{cage,imle,dreamvla,mdt} rely on the Markov assumption and predict future trajectories by conditioning only on short-term observations, thereby neglecting longer-term temporal dependencies.
However, state ambiguity is common in robotic manipulation\cite{sam2act,intro_SA_2,mtil}. Identical observations may correspond to multiple valid trajectories with the lack of contextual information. During data collection, human operators act as continuous reasoners with full awareness of the task context. 
For policy learning, many crucial cues for temporal disambiguation cannot be reflected from short observations.

Despite the importance of long history, it’s challenging to extend the history window in imitation learning. Naively extending the history window is computationally expensive\cite{tracevla,towards2412,contextvla} and may cause severe overfitting\cite{longdp}. Furthermore, manually crafted historical representations\cite{tracevla,hisaware} entail substantial annotation effort and struggle to handle diverse scenarios of state ambiguity.
The history window length is crucial. 
Some tasks, such as \textit{Wipe the table twice} (details in Figure \ref{fig:exp}), may require information from up to 7 seconds earlier to resolve state ambiguity. 
However, encoding long histories entails a trade-off between historical modeling capacity and inference cost, making stable and affordable training more challenging.

Moreover, different tasks require distinct types of historical information and varying lengths of history window. We identify five categories of scenarios illustrated in Figure \ref{fig:coarse}: (a) Bidirectional overlap patterns; (b) Fixed-duration static posture; (c) Sequential subtask dependencies; (d) Repetitive temporal cycles; (e) Dynamic interaction response. 
As shown in Figure \ref{fig:coarse}, take the \textit{hold the pot lid} as an example. During the back-and-forth motion, the robot must determine whether it should move upward or downward. When holding the pot lid, it needs to determine whether to continue holding it or to put it back. 

Human reasoning is inherently continuous rather than static, retaining the task context in working memory instead of discarding it. Meanwhile, working memory representations are not necessarily yoked to the sensory presentation format. Rather, our brain can flexibly recode information depending on task demands\cite{CongRepRecon,WMreconMIT}. 
Robotic long-history encoding can draw on human working memory in continuous reasoning processes by (1) forming adaptive working memory representations by recoding and (2) leveraging working memory to enable continuous reasoning over an extended history window, without processing all observations at each inference step.

Building on these insights, we introduce PAM, a novel visuomotor policy equipped with adaptive working memory. PAM allows a history window of up to 300 frames (the rationale for the length is detailed in Section \ref{history-window-length}).
During inference, a frame-level feature extractor fuses multimodal inputs, such as vision and language.
The module employs two types of query tokens: one extracts motion primitives directly relevant to action execution, while the other captures context features representing task progression and environmental changes, effectively serving as recoded working memory.
Next, a context router receives context features from historical inference steps and utilizes a set of query tokens spanning different history lengths to produce compact representation. 
Then it serves as supplementary context clues alongside the motion primitives to condition flow matching\cite{lipman2022flow} for predicting future action trajectories. Besides, humans recall past information when verification is needed. Thus a similar auxiliary objective is introduced in PAM to ensure the context router functions as an effective bottleneck.
Our contributions are as follows:

\begin{itemize}
\item[1)] 
Inspired by the continuous nature of human reasoning and the recoding of working memory, we propose PAM, a novel
visuomotor policy equipped with adaptive working memory. 
Adaptive recoding enables PAM to adapt effectively across diverse scenarios, without requiring additional annotations.

\item[2)]  PAM supports a long history window with a maximum length of 300 frames. It alleviates reliance on the Markov assumption and substantially extends the history window in a more efficient and stable manner, while preserving real-time inference at 20 Hz. Besides, PAM is interpretable, revealing which historical segments and modalities are used for disambiguation.

\item[3)] We carefully design 7 tasks covering different scenarios of state ambiguity. Compared to existing methods whose history windows are sufficient for these tasks, PAM achieves the best performance. 
In addition, PAM also achieves competitive performance on long-horizon tasks in Libero-Long.
 
\end{itemize}

\begin{figure*}[htp]
	\centering
	\includegraphics[width=1.0\linewidth]{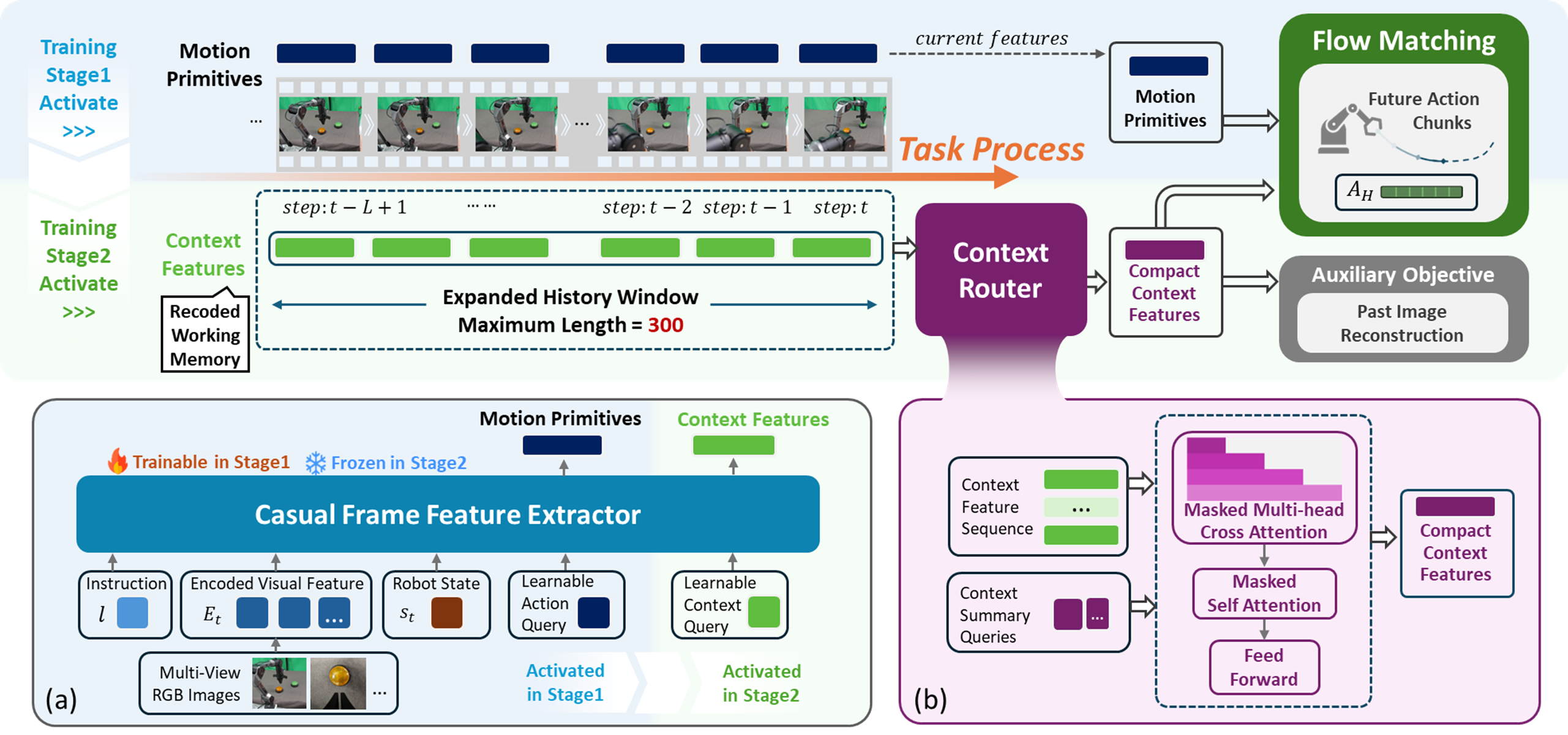}
	\captionsetup{font={small}}
	\caption{\textbf{Overview of PAM.} PAM uses two distinct types of features to guide action generation: \textcolor{primitives}{\textbf{motion primitives}} extracted from the current frame and \textcolor{compact}{\textbf{compact context features}} drawn from the extended history window. The \textcolor{context}{\textbf{context features}} serve as working memory. \textbf{(a)} illustrates the frame feature extractor used to obtain these features. The extractor employs a query-based mechanism to recode multimodal inputs, enabling adaptive working memory recoding. \textbf{(b)} illustrates the context router, which receives the the \textcolor{context}{\textbf{context features}} and utilizes a set of query tokens spanning different history lengths to produce \textcolor{compact}{\textbf{compact context features}}. PAM is trained in a two-stage manner. As shown in the figure, different subsets of model parameters are progressively activated across the two stages.
    } 
    \vspace{-0.5cm}
	\label{fig:pipeline}
\end{figure*}

\section{Related Work}
\label{sec:RelatedWork}

\subsection{History-Aware Visuomotor Policy}
History-aware visuomotor policies take various forms.
Many works are dedicated to establishing manual representations for long historical features. TraceVLA\cite{tracevla} provides rendered spatial traces that explicitly encode past motion. HistRISE \cite{hisaware} leverages object-centric tracking as additional input. Diff-control\cite{diff-control} and UniVLA\cite{univla} refer to historical actions when generating actions. These representations partially supplement missing clues but are limited in capacity and may incur high annotation costs.

Memory-bank approaches compress history into sparse tokens for storage. 4DVLA\cite{4dvla} performs dynamic sampling based on inter-frame differences. MemoryVLA\cite{memoryvla} designs retrieval, fusion, and consolidation mechanisms for memory bank. 
They preserve keyframes over long history windows but require complex design for selective memory storage, integration and retrieval. And it is challenging to design rules for encoding information like duration and repetition.

Another approach is multi-frame feature learning. ContextVLA\cite{contextvla} utilizes a vision-language model to understand image sequences. LongDP \cite{longdp} and CronusVLA \cite{cronusvla} use concatenated multi-frame perceptual features as denoising conditions, whereas UVA \cite{uva} first feeds them into a video generation model \cite{mar-uva}.
However, directly using features from a sliding window lacks adaptability and may cause feature confusion, while longer history windows increase costs and can hurt performance. PAM effectively enhances the adaptability of historical representations while mitigating the negative effects associated with longer history windows.

\subsection{Temporally Dependent Inference}

Recent advancements in many fields have increasingly adopted temporally dependent inference models\cite{mamba_s4,temporalcomprehensive,lian2025recent} that leverage temporal feature fusion or internal state evolution. HPNet\cite{tang2024hpnet} for vehicle trajectory prediction adopts historical prediction attention to enhance prediction consistency and accuracy. 
GhostShell\cite{ghostshell} exploits the streaming output of large language models to drive robotic actions in parallel.
OmniNav\cite{omninav} integrates a central memory module to support streaming inference in navigation.
Meanwhile, some approaches employ evolving internal states to dynamically model the environment. RoboVLMs\cite{robovlms} and RoboFlamingo\cite{roboflamingo} utilize memory-based LSTM networks to fuse temporal information across frames. MTIL\cite{mtil} encodes the full trajectory history by transmitting hidden states. These methods remain challenged by training cost and stability.

\section{Method}
\subsection{Overview of PAM}
\label{history-window-length}
\textbf{Problem Setup.} We aim to develop a visuomotor policy with a long history to resolve state ambiguity. Consider a multi-task dataset $\mathcal{D}$, where each task is specified by a language instruction $l$. Each task trajectory consists of observations $(o_{1}, \dots, o_{T})$, joint states $(\text{s}_{1}, \dots, \text{s}_{T})$, and executed actions $(a_{1}, \dots, a_{T})$. 
The variable $T$ represents the number of task frames. 
In a single inference step $t$, the policy outputs a sequence of future actions $A_t^H$, where $H$ is the length of the predicted action chunks.
PAM has a maximum history window length of 300 frames, while maintaining high-speed inference and stable training, which is unprecedented. 
The 300-frame history window covers the vast majority of scenarios requiring historical information (for details, see task analysis in Section \ref{tasks-7}). And scenarios that require longer context can be handled by the high-level task planner in robot system.

\textbf{Temporally Dependent Inference.} We reformulate policy inference as a temporally dependent process rather than as a set of independent steps. Most visuomotor policies operate with short history windows and predict in a static manner. At time step $t$, with a history window of length $h$:
\[
A^{H}_t= \mathcal{F}(l,o_{t-h}, \dots, o_{t},s_{t-h},\dots,s_t)
\]
At each time step $t$, PAM encodes only the current observation while maintaining the context features $c_t$ as working memory across inference steps.
Furthermore, the context features are leveraged in an auxiliary objective to predict embeddings of historical images.
\begin{equation}
A^H_t, \; c_t, \; E_t^{-L} = \mathcal{F}\big(l,o_t,s_t, \, \mathbf{C}_{t}^{\mathrm{hist}}\big)
\end{equation}
where the $E^{-L}_t$ is the historical image embeddings and \(\mathbf{C}_{t}^{\mathrm{hist}}\) denotes the set of context features from the history window.

\textbf{Adaptive Working Memory Recoding.} A key design decision in temporally dependent inference is determining which features should be propagated across inference steps, serving as working memory. Naively concatenating motion primitive features extracted from multimodal inputs introduces redundancy and produces contextual signals that lack essential contextual clues. 
Meanwhile, extracting working memory from the flow-matching-based action head is also suboptimal. In addition to limited representational capacity, the intermediate states generated over multiple denoising steps lack clear semantic grounding, making them unsuitable.
The human brain can flexibly recode information depending on task demands. Similarly, PAM employs a context query to recode information, forming an effective representation of working memory.
Enabled by causal attention, this context query can jointly attend to low-level perceptual cues and high-level decision-relevant features.

\subsection{Model Architecture}

\textbf{Casual Frame Feature Extractor.} 
In PAM, a DINOv2\cite{dinov2} image encoder with adapter layers is employed to encode raw RGB images
into a compact representation $e_t$.
The language instruction is encoded using a lightweight 6-layer version of MiniLM\cite{minilm} to obtain the embedding $l$. And a 3-layer MLP is adopted to encode the robot proprioception message as $s_t$.
The extractor employs a Transformer architecture with a causal mask and processes the encoded multimodal inputs as a sequence.
The input sequence is extended with two query tokens, $q^{act}$ and $q^{ctx}$. $q^{act}$ is used to encode the information for action primitives $e_t^{act}$. And as adaptive working memory recoding, the context query $q^{ctx}$ is used to encode the contextual features $c_t$. 

\textbf{Context Router.}
Unlike RNN-style approaches that propagate contextual features over time, the contextual features are fed into a context router as a sequence to produce compact representation. The context router is analogous to the neural systems that integrate and mobilize working-memory representations. This design effectively reduces feature confusion and overfitting, which is verified in ablation studies (Section \ref{ablation}). 
For the  set of context features 
\(\mathbf{C}_{t}^{\mathrm{hist}}\) from the history window,
a set of context-query tokens spanning different history lengths extracts key information from these features.
This design yields a hierarchical summary that captures both short-horizon and long-horizon information. For inference steps where the available history is shorter than the maximum history window, a masking strategy is applied by overlaying a validity mask on top of the original attention-span mask.
Then, as illustrated in Figure \ref{fig:pipeline}(c), the features are processed by a self-attention layer followed by a feed-forward layer, yielding the compact representation $e_t^{ctx}$. 

\textbf{Multi-conditioned Action Head.}
\label{headhead}
Conditioned on the action features $e_t^{act}$ and the compact context features $e_t^{ctx}$, the multi-conditioned action head predicts a sequence of action chunks based on flow matching. 
Specifically, the action head predicts the velocity field that transports an initial action state $A_t^{\text{noise}}$ toward the target action $A_t^H$. $A_t^{\text{interm}}$ denotes the intermediate feature produced during this flow-matching process.
The flow-matching objective supervises the predicted velocity $u_\theta$ at flow matching step $t^{\text{fm}}$ to align with this vector field:
\[
\mathcal{L}_{\mathrm{\text{FM}}}
= \mathbb{E}_{t^{\text{fm}}}
\left[
\left\|
u_\theta(A_t^{\text{interm}},\, e_t^{\text{act}}, e_t^{\text{ctx}},\, t^{\text{fm}}) - (A_t^H - A_t^{\text{noise}})
\right\|^2
\right]
\]

Four variants are considered for injecting multiple conditioning signals into the prediction network $u_\theta$: (1) Concatenate the two conditions; (2) Apply multi-head cross attention in a staggered manner, starting with motion primitives; (3) Apply multi-head cross attention in a staggered manner, starting with context features; (4)  A ControlNet-style multi-condition injection following the design in ControlVLA\cite{controlvla}.
The second variant is adopted as the default in our experiments, with corresponding ablation results presented in Section \ref{ablation}.

\textbf{Auxiliary Objective.}
Motivated by human recall processes, the module predicts past image features or trajectories, conditioned solely on $e_t^{\text{ctx}}$. 
The effectiveness of multimodal output has also been validated in previous works\cite{longdp,dreamvla}. Through early experiments, we found that predicting historical image embeddings is more effective in our settings.
Following the architecture in DINO-world\cite{dino-world}, we employ a decoder-only transformer architecture to predict past image embeddings. 
The patch features of historical observations are encoded using a VAE-based\cite{vae} autoencoder, yielding the sequence $E^{-L}_t$, where $L$ is the sequence length. The auxiliary objective is defined as: 
\[
\mathcal{L}_{\mathrm{aux,img}} = \| \hat{E}^{-L}_t - E^{-L}_t \|_2^2.
\]

\subsection{Training Recipe}
Temporally dependent inference modeling presents significant challenges for the training recipe. PAM is trained in a two-stage manner and maintains efficient parallel sampling in both stages. It avoids the instability and high training cost associated with training temporally dependent inference models in a sequential way\cite{mtil,hpnet}. And this strategy is scalable to larger models.

\textbf{Stage 1: Learning Motion Primitives} As shown in Figure \ref{fig:pipeline}(a), in the first stage, the action head directly takes the motion primitives $e^{act}$ and predicts future action chunks. The parameters related to $e^{ctx}$ are not activated. 
When encountering state ambiguity, the predicted trajectories often fluctuate across multiple modes. This stage aims to train a stable multimodal frame extractor in preparation for the subsequent stage. 
For tasks without pronounced state ambiguity, the policy is already capable of completing them effectively.

\textbf{Stage 2: Temporal Disambiguation} After the first stage, the extractor is frozen, and the key–value states are cached for subsequent training, as illustrated in Figure \ref{fig:pipeline}(b). Consequently, this stage no longer requires raw perceptual inputs and maintains efficient parallel sampling.
The stage enables PAM to develop robust temporal disambiguation capabilities.
The overall loss is:
\[
\mathcal{L}
= \mathcal{L}_{\mathrm{FM}}
+ \lambda_{\mathrm{aux}}\ \mathcal{L}_{\mathrm{aux}},
\]
where $\lambda_{\mathrm{aux}}$ balances action learning and contextual verification.  $\lambda_{\mathrm{aux}}$ is set to 0.4. 

\begin{figure*}[htp]
	\centering
	\includegraphics[width=1.0\linewidth]{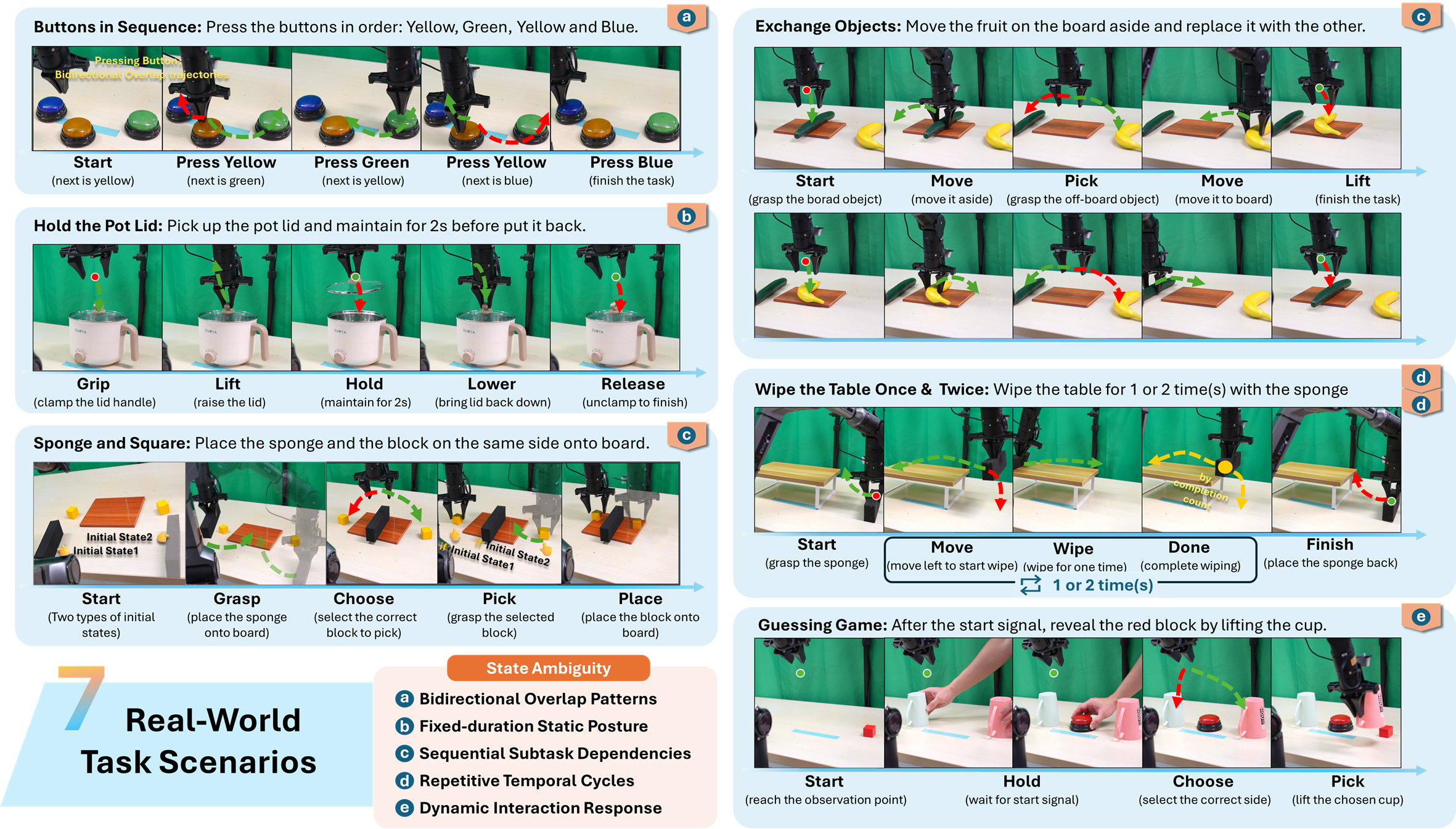}
	\captionsetup{font={small}}
	\caption{\textbf{Real-world Tasks.} We carefully designed a set of real-world tasks as shown in the figure, with the primary types of state ambiguity indicated in the top-right corner. Each task comprises multiple subtasks, and to provide a more fine-grained evaluation, the task success rate is computed as the average completion rate across its subtasks.}
        \vspace{-0.5cm}
	\label{fig:exp}
\end{figure*}

\section{Experiments}
In the following sections, we first evaluate PAM’s ability to resolve state ambiguity in real-world tasks (Sections \ref{setting-real} and \ref{sec:real-world}). We then assess its performance on Libero-Long\cite{libero} long-horizon tasks (Section \ref{sec:sim}). Lastly, we perform ablation studies to examine the efficacy of each module and the impact of various parameter configurations on overall performance (Section \ref{ablation}).

\subsection{Real-World Experimental Setting}
\label{setting-real}
\textbf{Platform.}
For our real-world experiments, we employ an Agilex Piper, a six-axis robotic arm equipped with a parallel gripper. The visual input consists of RGB images at a resolution of 320x240, captured a fixed global camera and a wrist camera.
The action space is defined by the robot's joint angles and the gripper’s state.
For each task, we collected 50 expert trajectories at a frequency of 30 Hz using a master-slaver teleoperation setup.

\textbf{Task Settings.}
\label{tasks-7}
We design a suite of tasks to evaluate the model’s capability under diverse state ambiguity scenarios. The task procedures for the 7 tasks are illustrated in Figure \ref{fig:exp}. 
Two representative tasks are \textit{Wipe Two Times} and \textit{Guessing Game}.
In \textit{Wipe Two Times}, the end effector passes through the right side of the table three times, each marking a distinct task phase: (i) initiating the first wipe after picking up the sponge, (ii) starting the second wipe after the first is completed, and (iii) returning the sponge to its original position. Approximately 210 frames elapse between the first and third passes. 
Since the two wiping passes follow identical trajectories, the robot must consider the full observation history; otherwise, it is prone to stereotypical behaviors, such as redundant wiping, premature returning, or decision instability. In \textit{Guessing Game}, the robot must maintain its working memory during a significant delay; approximately 100 frames elapse from the moment the block positions are occluded until the robot is permitted to act. 
Distinct from the prior task, it relies on passive observation rather than active interaction for information acquisition.
These two tasks exemplify scenarios that require a long history window, which is fully supported by PAM with a maximum length of 300 frames.

\textbf{Baselines.}
We choose two representative history aware visuomotor policies in real-world experiments: MTIL\cite{mtil} and the LongDP\cite{longdp}. Like PAM, they also need no manual annotations and are trained from scratch. 
LongDP enhances long-horizon temporal modeling by incorporating a self-supervised objective to reconstruct past action tokens. 
During inference, LongDP combines multi-frame perceptual representation to perform action decoding.
MTIL uses a selective state-space model to encode full trajectory history into a compact latent state, allowing the policy to be conditioned on a comprehensive temporal context.

\textbf{Implementations and metrics.}
For real-world evaluation, all policies are trained on an A100 GPU and depolyed on an RTX4060 GPU. The inference frequency of PAM is set at 20Hz. For MTIL and LongDP, we have determined their maximum feasible frequencies as 20Hz and 5Hz. Both PAM and MTIL adpot the temporal aggregation strategies\cite{ACT}. And for LongDP, a self-verification step selects action samples most consistent with predicted trajectories. PAM is trained as a unified multi-task policy for all scenarios in both real-world and simulation environments. LongDP and MTIL follow a single-task training paradigm, where separate models are optimized for each task. Apart from the image and language encoders, all parameters of these models are trained from scratch.
Each policy is evaluated over 20 independent trials per task. Given that our tasks typically comprise 2–3 sequential sub-tasks, we report the overall success rate as the mean success rate across these internal stages, averaged over the 20 trials.

\begin{table*}[htp]
\centering
\renewcommand{\arraystretch}{1.2}
\caption{Performance comparison across diverse real-world tasks.}
\begin{tabular}{l M{1.1cm} M{1.3cm} M{1.2cm} M{1.4cm} M{1.2cm} M{1.4cm} M{1.5cm} M{1.0cm} M{1.1cm}}
\toprule
\textbf{Method} & \textbf{History Window} & \textbf{Buttons in Sequence} & \textbf{Hold the Pot Lid} & \textbf{Sponge and Square} & \textbf{Exchange   Objects} & \textbf{Wipe the Table Once} & \textbf{Wipe the Table Twice} & \textbf{Guessing   Game} & \textbf{Average} \\
\midrule
LongDP & 300 & 0.51 & 0.55 & 0.77 & 0.62 & 0.70 & 0.56 & 0.57 & 0.61 \\
MTIL &$\infty$ &  0.45 & 0.20 & 0.60 & 0.44 & 0.50 & 0.35 & 0.60 & 0.45 \\

\cellcolor{lightours}PAM (ours) & \cellcolor{lightours}300 
& \cellcolor{lightours}\textbf{0.80} 
& \cellcolor{lightours}\textbf{0.95} 
& \cellcolor{lightours}\textbf{0.96} 
& \cellcolor{lightours}\textbf{0.90} 
& \cellcolor{lightours}\textbf{1.00} 
& \cellcolor{lightours}0.89 
& \cellcolor{lightours}0.90 
& \cellcolor{lightours}\textbf{0.91} \\

\cellcolor{lightours}PAM w/ aux. head & \cellcolor{lightours}300 
& \cellcolor{lightours}0.76 
& \cellcolor{lightours}\textbf{0.95} 
& \cellcolor{lightours}0.95 
& \cellcolor{lightours}0.80 
& \cellcolor{lightours}\textbf{1.00} 
& \cellcolor{lightours}\textbf{0.92} 
& \cellcolor{lightours}\textbf{0.97} 
& \cellcolor{lightours}\textbf{0.91} \\

\bottomrule
\end{tabular}
\label{tab:result}
\end{table*}

\begin{figure*}[htp]
	\centering
	\includegraphics[width=0.97\linewidth]{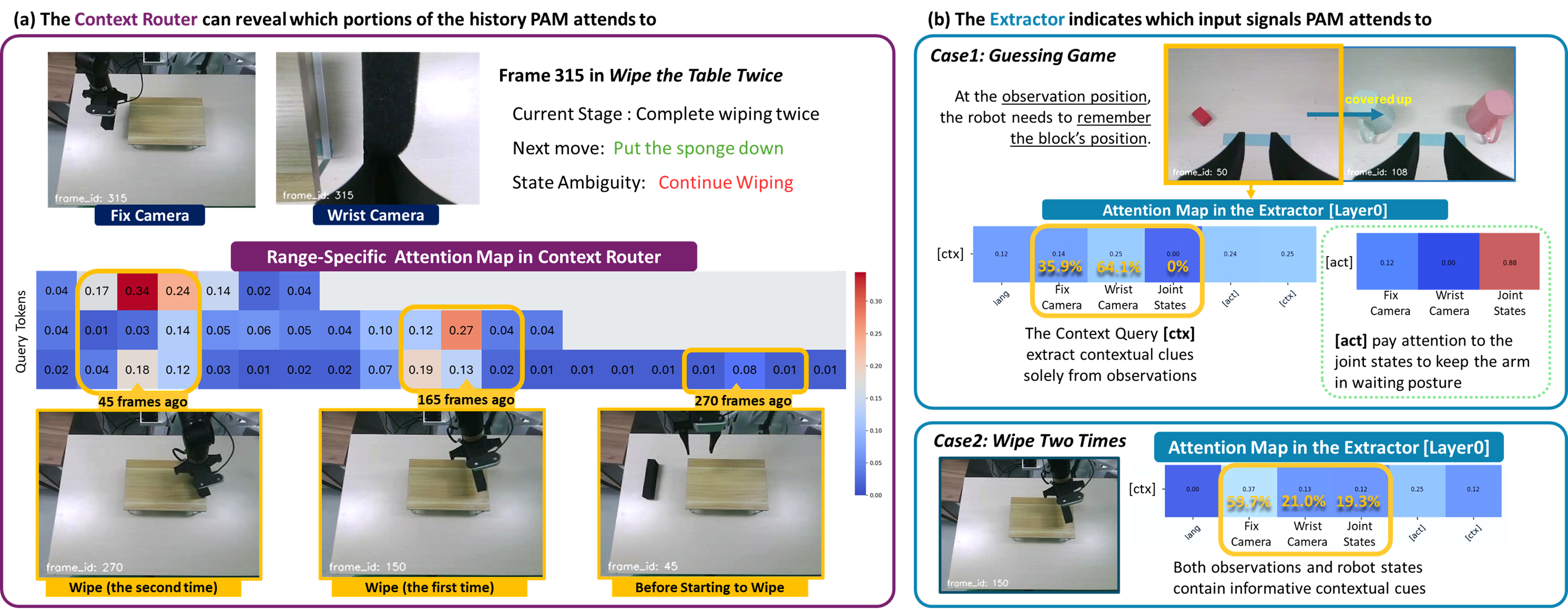}
	\captionsetup{font={small}}
	\caption{PAM exhibits strong interpretability. \textbf{Left:} The attention maps of the context router reveal which history portions PAM references to resolve state ambiguity, accurately identifying key frames from preceding task stages in \textit{Wipe the Table Twice}. \textbf{Right:} The attention maps of extractor indicate which modalities PAM uses for working memory encoding. In \textit{Guessing Game}, the context query extracts historical cues from visual observation of the block's position, while the motion primitives attends to joint states for posture maintenace. In \textit{Wipe the Table Twice}, both visual observations and joint states provide effective contextual cues, demonstrating the effectiveness of our adaptive working memory encoding.
    }
        \vspace{-0.5cm}
	\label{fig:attn}
\end{figure*}

\subsection{Main Results in Real-World Tasks}
\label{sec:real-world}

Table \ref{tab:result} reports the experimental results. PAM demonstrates strong performance across all tasks requiring diverse contextual cues. Notably, PAM maintains high and consistent success rates (averaging 0.91) across various scenarios. This represents a significant performance gain of 49.2\% over the LongDP (0.61) baseline and an over twofold improvement (102.2\%) compared to MTIL (0.45). 
PAM's advantages are particularly pronounced in several challenging scenarios. For instance, in the \textit{Hold the Pot Lid} and \textit{Wipe the Table Twice} tasks, the baselines encounter substantial difficulties; in contrast, PAM achieves success rates of 0.95 and 0.92, a performance gap that is largely attributable to PAM’s ability to learn highly discriminative history representations.

LongDP achieves strong performance on several tasks, such as \textit{Sponge and Square} and \textit{Wipe One Time}. However, because it directly conditions its denoising process on raw multi-frame perceptual features, it frequently encounters perceptual confusion in multi-stage tasks like \textit{Wipe Two Times}. Furthermore, this high-dimensional conditioning leads to significant computational overhead, resulting in slower inference.
MTIL employs a forward-propagated hidden state as its working memory and is theoretically capable of supporting an unbounded history window. 
In practice, however, this hidden state is highly susceptible to representational distortion. This often manifests as impulsive or erratic actions during the initial task phases. And the robot exhibits pronounced jitter and jittery trajectories during the initial stages of a task.
To overcome these challenges, PAM employs adaptive recoding to distill compact, informative representations for temporal disambiguation. Its ability to perform seven tasks simultaneously using shared parameters demonstrates the efficacy of the context router in distilling task-relevant compact representations.

Moreover, PAM’s overall performance is largely consistent with or without the auxiliary prediction head. The auxiliary objective is disabled during inference, thus the inference cost remains identical. However, its impact varies across tasks. In \textit{Guessing Game}, the auxiliary head yields a marked improvement in success rate. We attribute this to the fact that predicting historical image embeddings encourages more focused attention on early visual cues and thereby stabilizes the spatial memory of the cube’s initial location. At the same time, we find that assigning excessively large weights to the auxiliary objective can reduce the smoothness of the arm’s motion. 

Beyond performance, PAM exhibits strong interpretability through its structured representational space. As illustrated in Figure \ref{fig:attn}, the attention maps of the context router and the extractor visualize the policy's decision-making logic, revealing  which historical segments and sensory modalities are leveraged to resolve state ambiguity.

\subsection{Evaluation on Libero-Long Long-Horizon Tasks}
\label{sec:sim}
Beyond its capacity to resolve state ambiguity, PAM demonstrates robust performance across a diverse range of manipulation challenges.
Libero-Long\cite{libero} provides a procedurally generated suite of long-horizon robotic manipulation tasks. We evaluated PAM’s performance on 10 tasks from Libero-Long to assess its capabilities in long-horizon manipulation.

We utilize the official LIBERO-Long dataset, with each task consisting of 50 expert trajectories. To benchmark PAM’s performance, we compare it against several strong baselines: OpenVLA\cite{openvla}, Diffusion Policy(DP)\cite{dp}, MDT\cite{mdt}, and $\pi_0$\cite{pi0}. The default image size of 128×128 is used for learning. Given that these tasks do not exhibit pronounced state ambiguity, the history window of PAM is set to 50.
For evaluation, each task is tested across five random seeds, totaling 100 trials per task. We report the average success rate in Table \ref{tab:sim}.

\begin{table}[!htbp]
\centering
\caption{Results on the Libero-Long 10 tasks.}
\begin{tabular}{c c c c c c}
\toprule
\textbf{Method} & OpenVLA & DP-T & MDT & $\pi_0$ &
\begin{tabular}{@{}c@{}}{PAM} \\ (ours)\end{tabular} \\
\midrule
\textbf{\begin{tabular}{@{}c@{}}Avg. Success Rate\end{tabular}} 
& 0.544 & 0.582 & 0.653 & 0.852 & 0.847 \\
\bottomrule
\end{tabular}
\label{tab:sim}
\end{table}

PAM reaches an 84.7\% average success rate, matching $\pi_0$ and significantly exceeding other baselines. This demonstrates that PAM's architecture, though designed for state disambiguation, generalizes effectively to long-horizon manipulation tasks.

\subsection{Ablation Studies}
\label{ablation}
Our ablation studies focus on 4 critical factors: (1) visual encoder selection; (2) multi-condition injection mechanism (details in Section \ref{headhead}); (3) variants of the context router; (4) sampling interval for the context features. The configuration we ultimately adopt is highlighted, corresponding to the setup in \ref{sec:real-world}.

\textbf{Visual Encoder.} As shown in Table \ref{tab:ablationtable}, employing a lightweight DINOv2 encoder\cite{dinov2} leads to a substantial performance degradation. Empirical observations suggest that the policy exhibits more stereotypical behaviors in this configuration. 
For instance, in \textit{Guessing Game}, the policy tends to bias its predictions toward a specific direction. 
This is because the history representation must have access to sufficient information to extract relevant contextual cues for temporal disambiguation.
In this case, the red cube's low pixel occupancy renders it susceptible to insufficient representation or perceptual confusion when processed by a lightweight encoder.

\begin{table}[!htbp]
\centering
\caption{Success rates in ablation studies}
\renewcommand{\arraystretch}{1.2}

\begin{tabularx}{0.9\linewidth}{c X c}
\toprule
 & \textbf{Variant} & \textbf{Average} \\ \midrule

\multirow{2}{*}{\centering \textbf{Visual Encoder}} 
 & \cellcolor{lightgreen}ViT-Base 
 & \cellcolor{lightgreen}\textbf{0.91} \\
 & ViT-Small & 0.72 \\ \midrule

\multirow{4}{*}{\centering \textbf{Condition Type}} 
 & \cellcolor{lightgreen}Later Inject Context Features
 & \cellcolor{lightgreen}\textbf{0.91 }\\
 & Parallel Inject & 0.89 \\
 & First Inject Context Features& 0.72 \\
 & Concatenate & 0.69 \\ \midrule

\multirow{5}{*}{\centering \textbf{Context Router}} 
 & Avg Token (No Router) & 0.59 \\
 & 1 Query Token & 0.75 \\
 & \cellcolor{lightgreen}3 Query Tokens
 & \cellcolor{lightgreen}\textbf{0.91} \\
 & 5 Query Tokens & 0.80 \\
 & 3 Query Tokens + Avg Token & 0.71 \\ \midrule

\multirow{5}{*}{\centering \textbf{Sample interval}} 
 & 5 & 0.81 \\
 & 10 & 0.81 \\
 & \cellcolor{lightgreen}15 
 & \cellcolor{lightgreen}\textbf{0.91} \\
 & 20 & 0.86 \\
 & 25 & 0.83 \\

\bottomrule
\end{tabularx}

\label{tab:ablationtable}
\end{table}

\textbf{Condition Type.} 
While recoding the working memory is essential, the mechanism for deploying it alongside motion primitives to guide action generation is equally critical.
In PAM, the two conditions have clear roles: one provides motion primitives, and the other supplies contextual cues. According to results in Table \ref{tab:ablationtable}, the most effective variants are either sequential cross-attention (applying $e^{act}$ followed by $e^{ctx}$) or incorporating the working memory via a ControlNet-style adapter. 
We found that early injection of motion primitives is essential for guiding the action manifold transition. Suboptimal results from the other two variants suggest that an improper conditional order or method fails to provide the structured guidance necessary for coherent action decoding.

\textbf{Context Router.} As shown in Table \ref{tab:ablationtable}, 
replacing the context router with a simple average pooling of all context features yielded the poorest performance. 
Since the query tokens attend to different history lengths in a staged manner, using too few query tokens fails to effectively integrate contextual cues, which are often distributed across multiple historical frames. Conversely, employing too many query tokens results in less compact representations, imposing additional burden on action decoding.

\textbf{Sample Interval.} Considering the minimal changes between adjacent frames and the duration of visual persistence, we adopt interval sampling as a trade-off between computational cost and information density. 
An overly small sampling interval leads to highly redundant representations, hindering the learning of compact and effective features. Conversely, an excessively large interval reduces information density, potentially missing critical contextual cues and increasing the policy’s sensitivity to distribution shifts during inference.

\section{Conclusion}
\label{sec:Conclusion}
In this work, we introduce PAM, a  visuomotor policy equipped with adaptive working memory and a 300-frame history window. By combining the training scheme with compatible model architectures, It supports efficient parallel sampling to develop the ability of temporal disambiguation. We provide a clear and practical way to obtain a long history window with affordable cost while maintaining high inference speed.
PAM demonstrates strong performance on tasks requiring diverse contextual cues while also exhibiting great versatility. It also has good performance in long-horizon tasks. Furthermore, PAM’s architecture and design principles are simple, facilitating extension to larger models.

\newlength{\bibitemsep}\setlength{\bibitemsep}{0.00\baselineskip}
\newlength{\bibparskip}\setlength{\bibparskip}{0pt}
\let\oldthebibliography\thebibliography
\renewcommand\thebibliography[1]{
	\oldthebibliography{#1}
	\setlength{\parskip}{\bibitemsep}
	\setlength{\itemsep}{\bibparskip}
}
\bibliography{arxiv}

\end{document}